\documentclass{article}
\usepackage{ijcai19}

\usepackage{times}
\usepackage{soul}
\usepackage{url}
\usepackage[hidelinks]{hyperref}
\usepackage[utf8]{inputenc}
\usepackage[small]{caption}
\usepackage{graphicx}
\usepackage{amsmath}
\usepackage{booktabs,bm}
\usepackage{algorithm}
\usepackage{algorithmic}

\usepackage{epsfig}
\usepackage{amssymb}
\usepackage{amsfonts}
\usepackage{textcomp}
\usepackage{xcolor}
\usepackage{color}
\usepackage{dirtytalk}
\usepackage{subfigure}
\usepackage{multirow}
\usepackage{epstopdf}
\usepackage{setspace}
\usepackage{enumitem}
\usepackage{float}
\usepackage{hyperref}
\usepackage{capt-of}

\title{Protecting Neural Networks with Hierarchical Random Switching: Towards Better Robustness-Accuracy Trade-off for Stochastic Defenses}

\author{
Xiao Wang$^1$\footnote{Equal Contribution}\addtocounter{footnote}{-1}\addtocounter{Hfootnote}{-1}\and
Siyue Wang$^2$\footnotemark\and
Pin-Yu Chen$^3$\and
Yanzhi Wang$^2$\and\\
Brian Kulis$^1$\and
Xue Lin$^2$\And
Peter Chin$^1$
\\
\affiliations
$^1$Boston University \\
$^2$Northeastern University \\
$^3$IBM Research\\
}

\begin{document}

\maketitle

\begin{abstract}

Despite achieving remarkable success in various domains, recent studies have uncovered the vulnerability of deep neural networks to adversarial perturbations, creating concerns on model generalizability and
new threats such as prediction-evasive misclassification or stealthy reprogramming.
Among different defense proposals, stochastic network defenses such as random neuron activation pruning or random  perturbation to layer inputs are shown to be promising for attack mitigation. However, one critical drawback of current defenses is that the robustness enhancement is at the cost of noticeable performance degradation on legitimate data, e.g., large drop in test accuracy. This paper is motivated by pursuing for a better trade-off between adversarial robustness and test accuracy for stochastic network defenses. We propose Defense Efficiency Score (DES), a comprehensive metric that measures the gain in unsuccessful attack attempts at the cost of drop in test accuracy of any defense. To achieve a better DES, we propose hierarchical random switching (HRS), which protects neural networks through a novel randomization scheme. A HRS-protected model contains several blocks of randomly switching channels to prevent adversaries from exploiting fixed model structures and parameters for their malicious purposes. Extensive experiments show that HRS is superior in defending against state-of-the-art white-box and adaptive adversarial misclassification attacks. We also demonstrate the effectiveness of HRS in defending adversarial reprogramming, which is the first defense against adversarial programs. Moreover, in most settings the average DES of HRS is at least 5$\times$ higher than current stochastic network defenses, validating its significantly improved robustness-accuracy trade-off.

\end{abstract}

\section{Introduction}
\label{intro}





Deep neural networks (DNNs) have led to substantial improvements in the field of computer vision \cite{lecun1998gradient,wang2018using}, natural language processing \cite{hu2014convolutional} and automatic decision making \cite{mazurowski2008training}, and have influenced a broad range of real-world applications. Nonetheless, even under a simple norm-ball based input perturbation threat model,
they are recently shown to struggle with adversarial examples such as adversarial misclassification attacks \cite{szegedy2013intriguing,Goodfellow2015explaining,carlini2017adversarial,su2018robustness,zhao2018admm}, or adversarial reprogramming \cite{elsayed2018adversarial}, bringing about increasing concerns on model generalizability and new security threats \cite{Zhao2019fault}.

Among different defense proposals, stochastic network defenses
are shown to be promising for mitigating adversarial effects. The key idea is to replace a deterministic model with a stochastic one, with some parameters being randomized. Popular stochastic network defenses include stochastic activation pruning (SAP) \cite{s.2018stochastic}, defensive dropout \cite{wang2018defensive} and adding Gaussian noise \cite{liu2017towards}. The variation of a stochastic model leads to stochastic input gradients and therefore perplexes the adversary when crafting adversarial examples. However, one critical drawback of current defenses is the noticeable drop in model performance on legitimate data (particularly, the test accuracy), resulting in an undesirable trade-off between defense effectiveness and test accuracy. 


In pursuit of a better trade-off between defense effectiveness and test accuracy, in this paper we propose a novel randomization scheme called  hierarchical random switching (HRS). A HRS-protected network is made of a chain of random switching blocks. Each block contains a bunch of parallel channels with different weights and a random switcher controlling which channel to be activated for taking the block's input.
In the run time, the input is propagated through only the activated channel of each block and the active channels are ever-switching. Note that each activated path in HRS-protect model features decentralized randomization for improved robustness but is also fully functional for legitimate data, which is expected to yield a better trade-off. In addition, different from the ensemble defense using multiple different networks \cite{tramer2017ensemble}, HRS only requires one single base network architecture to launch defense.

To rigorously evaluate the adversarial robustness of the proposed HRS method, in Section \ref{sec_performance} we adopt the security inspection principles suggested in \cite{athalye2018obfuscated} to verify the robustness claims\footnote{\label{ref_code_appen}Appendices and codes: \url{https://github.com/KieranXWang/HRS}}.
Specifically, we mount four widely used adversarial misclassification attack methods (FGSM, CW, PGD and CW-PGD)  under different scenarios, including standard white-box and two adaptive white-box attack settings. The adaptive attacks  consider the setting where the adversary has the additional knowledge of randomization being deployed as defenses.
The results show the superior defense performance of HRS, and most importantly, ensuring its robustness is indeed NOT caused by ``security through obscurity'' such as gradient obfuscation \cite{athalye2018obfuscated}. 

 Below we summarize our main contributions.
\begin{itemize}[leftmargin=*]
    \item We propose a novel metric called \textit{defense efficiency score} (DES) that evaluates the trade-off between test accuracy and defense rate on adversarial examples, which provides a standardized approach to compare different defenses.  In particular, 
    we analyze the trade-off in state-of-the-art stochastic network defenses and adversarial training to motivate DES in Section \ref{sec_DEI}.
    
    \item To achieve a better robustness-accuracy trade-off, in Section \ref{sec_HRS} we propose hierarchical random switching (HRS), an easily configurable stochastic network defense method that achieves significantly higher DES than current methods.
    HRS is an attack-independent defense method that can be easily mounted on the typical neural network training pipelines. 
    We also develope a novel bottom-up training algorithm with linear time complexity that can effectively train a HRS-protected network. 
    
    \item Compared with state-of-the-art stochastic defense methods (SAP, defensive dropout and Gaussian noise), experiments on MNIST and CIFAR-10  show that HRS exhibits much stronger resiliency to adversarial attacks and simultaneously sacrifices less test accuracy. Moreover,  HRS is an effective defense against the considered powerful adaptive attacks that break other stochastic network defenses, which can be explained by its  unique decentralized randomization feature (see Section \ref{sec_HRS} for details). 
    
    \item HRS can effectively mitigate different adversarial threats. In Section \ref{sec_adv_reprog}, we show that HRS is an effective defense against adversarial reprogramming \cite{elsayed2018adversarial}. To the best of our knowledge, this paper proposes the first defense against adversarial reprogramming. 
    

\end{itemize}

\section{Adversarial Threats}
\subsection{Adversarial Misclassification Attack}
Fast Gradient Sign Method (\textbf{FGSM}) \cite{Goodfellow2015explaining} is a \say{one-shot} attack that generates an adversarial example $x'$ by taking one step gradient update in the $\ell_\infty$ neighborhood of input image $x$ with a step size $\epsilon$. 



 Carlini \& Wagner (\textbf{CW}) attack \cite{carlini2017towards} formulates the search for adversarial examples by solving the following optimization problem:
\begin{equation}
\text{minimize}_{\delta} \quad D(\delta) + c\cdot f (x+\delta) \quad \text{s.t.} \quad x+\delta \in [0,1]^n
\label{cw object}
\end{equation}
where $\delta$ denotes the perturbation to $x$. $D(\delta)$ is the distortion metric; $f$ is a designed attack objective for misclassification;
and the optimal term $c>0$ is obtained by binary search. 

Projected Gradient Decent  (\textbf{PGD})  \cite{madry2017towards} is an iterative attack that applies FGSM with a small step size $\alpha$. It controls the distortion of adversarial examples through clipping the updated image so that the new image stays in the $\epsilon$ neighborhood of $x$. For the $t$-th iteration, the adversarial image generation process is:
\begin{equation}
x'_{t+1} = \prod_{x+S}( x'_t - \alpha \cdot \text{sign} (\nabla(loss_{t}(x))))
 \label{sap}
\end{equation}
where $\prod_{x+S}$ means projection to $S$, the allowed perturbation in an $\ell_\infty$ ball centered at $x$, \text{sign} applies element-wise, and $\nabla loss(\cdot)$ is the gradient of misclassification loss.

\textbf{CW-PGD} \cite{athalye2018obfuscated} applies the loss term $f$ in CW attack to PGD attack. Different from CW attack, CW-PGD can directly control the level of distortion by clipping updated images into an $l_\infty$ ball of radius $\epsilon$.




\subsection{Adversarial Reprogramming}
Adversarial reprogramming \cite{elsayed2018adversarial} is a recent adversarial threat that aims at \say{reprogramming} a target model trained on task $T_a$ into performing another task $T_b$. It is accomplished by learning an input transformation $h_f$ and an output transformation $h_g$ that bridge the inputs and outputs of $T_a$ and $T_b$. After reprogramming, the computational cost of performing task $T_b$ only depends on $h_f$ and $h_g$, so that the attacker can stealthily exploit the target model.

\section{Stochastic Network Defenses: Motivation and Background}
Here we provide some motivation and background on why and how randomness can be exploited to defend against adversarial attacks. We are particularly interested in stochastic network defenses due to the following reasons: (i) they exhibit promising robustness performance; (ii) they are easily compatible with typical neural network training procedures; and (iii) they do not depend on specific adversarial attacks for training robust models, such as adversarial training \cite{Goodfellow2015explaining,madry2017towards}.

\subsection{Motivation}
Why randomness can be useful in defending adversarial attacks? Here we conduct two sets of experiments and report some insightful observations. First, we find that following a randomly selected direction, the possibility  of finding a successful adversarial example is very low. Second, we find that when training models with the same network architecture but with different weight initialization, each model has its own vulnerable spots. Details of these experiments can be found in Appendix\footnotemark[1] \ref{sup_motivation}.

Combining these two findings, one can reach an intuitive motivation on why stochastic network defenses can be effective. As adversarial attacks are associated with worst-case performance and the model vulnerability varies with initial weight randomization, a successful attack on a stochastic model requires finding a common weakness that applies to ALL stochastic model variants. This indicates that finding an effective adversarial example for a randomized network is strictly more difficult than that for a deterministic model.

\subsection{Background}
There are many defense methods that utilize randomness either explicitly or implicitly. Here we summarize three representative works toward this direction.

\paragraph{Stochastic Activation Pruning (SAP).}
Stochastic activation pruning (SAP), proposed by Dhillon et al. \cite{s.2018stochastic}, randomizes the neural network by stochastically dropping neuron outputs with a weighted probability. After pruning, the remaining neuron outputs are properly scaled up according to the number of pruned neurons. 



\paragraph{Defensive Dropout.}
Wang et al. \cite{wang2018defensive} propose defensive dropout that applies dropout \cite{srivastava2014dropout} in the inference phase for defense. Defensive dropout differs from SAP in two aspects. First, it drops neurons equally regardless of their magnitudes. Second, defensive dropout is implemented in both training and testing phases, with possibly different rates. 


\paragraph{Gaussian Noise.}
Liu et al. \cite{liu2017towards} introduce randomness to the network by adding Gaussian noise before each convolutional layer. Similar to defensive dropout, Gaussian noise takes place in both training and testing phases. The authors suggest to use different noise deviations for the input convolutional layer and other convolutioanl layers, which they refer to as "init-noise" and "inner-noise" respectively.


\subsection{Stochastic Gradients}




\begin{table}[t]
    \centering
    \scalebox{0.9}{
    \begin{tabular}{|c|c|c|}
        \hline
         Model & Deviation  & Defense Rate (\%) \\
        \hline
        \hline
        Base & 0 & 29.18 \\
        SAP & 0.0343 & 29.40\\
        Dropout 0.1 & 0.1143 & 29.60\\
        Dropout 0.3 & 0.2295 & 29.63\\
        Dropout 0.7 & 0.5186 & 32.62\\
        Gaussian & 0.6413 & 39.92\\
        HRS $10\times10$ & 0.7376 & 61.03\\
        HRS $20\times20$ & 0.7888 & 66.67\\
        HRS $30\times30$ & 0.7983 & 69.70\\
        \hline
    \end{tabular}}
    \label{gradient_dev_tab}
    \caption{Input gradient standard deviation and mean defense rate (1 - attack success rate) under PGD attack of multiple strengths. Details are in Section \ref{sec_performance}. A visualization plot is given in Appendix \ref{sup_deviation}.}
\end{table}

\begin{figure}[t]
    \centering
    \includegraphics[width=0.48\textwidth]{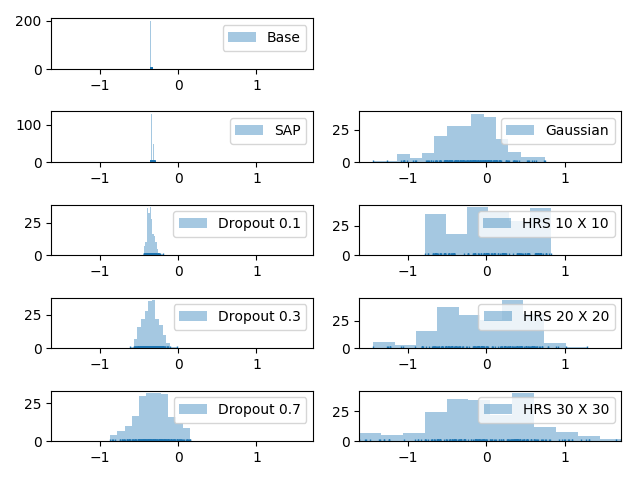}
    \caption{An example of input gradient distribution of stochastic defense models on a randomly selected dimension. We sample the input gradient of each defense for 200 times at the first step of CW-PGD attack. While SAP, dropout and Gaussian noise all yield a unimodal distribution, this trend is less obvious for HRS.}
    \label{fig:gradient_dist_sample}
\end{figure}

One unique property of stochastic models is the consequence of stochastic input gradients when performing backpropagation from the model output to the input. By inspecting the mean standard deviation of input gradient under different attack strengths ($\ell_\infty$ constraint) over each input dimension, we find that it is strongly  correlated with the defense performance, as shown in Table \ref{gradient_dev_tab}. By simply mounting a white-box attack (PGD), we find that SAP becomes as vulnerable as the base (deterministic) model, and defensive dropout and Gaussian noise have little defense effects. However, defenses that have larger standard deviations of the input gradient, such as the proposed HRS method (see Section \ref{sec_HRS} for details), are still quite resilient to this white-box attack.
For visual comparison, an example of different stochastic models' input gradient distributions under the same attack is illustrated in Figure \ref{fig:gradient_dist_sample}.


\section{Defense Efficiency Score: Quantifying Robustness-Accuracy Trade-off}
\label{sec_DEI}

For most current defense methods, there are factors controlling the defense strength. For example, in adversarial training \cite{madry2017towards}, one can achieve different defense strength by using different $L_\infty$ bounds on adversarial perturbations during training. Analogously, for stochastic defenses, the controlling factors are the randomization sources, such as the dropout rate, variance of Gaussian noise or the width (number of channels) in our proposed HRS approach. 

\begin{figure}[t]
    \centering
    \includegraphics[width=0.48\textwidth]{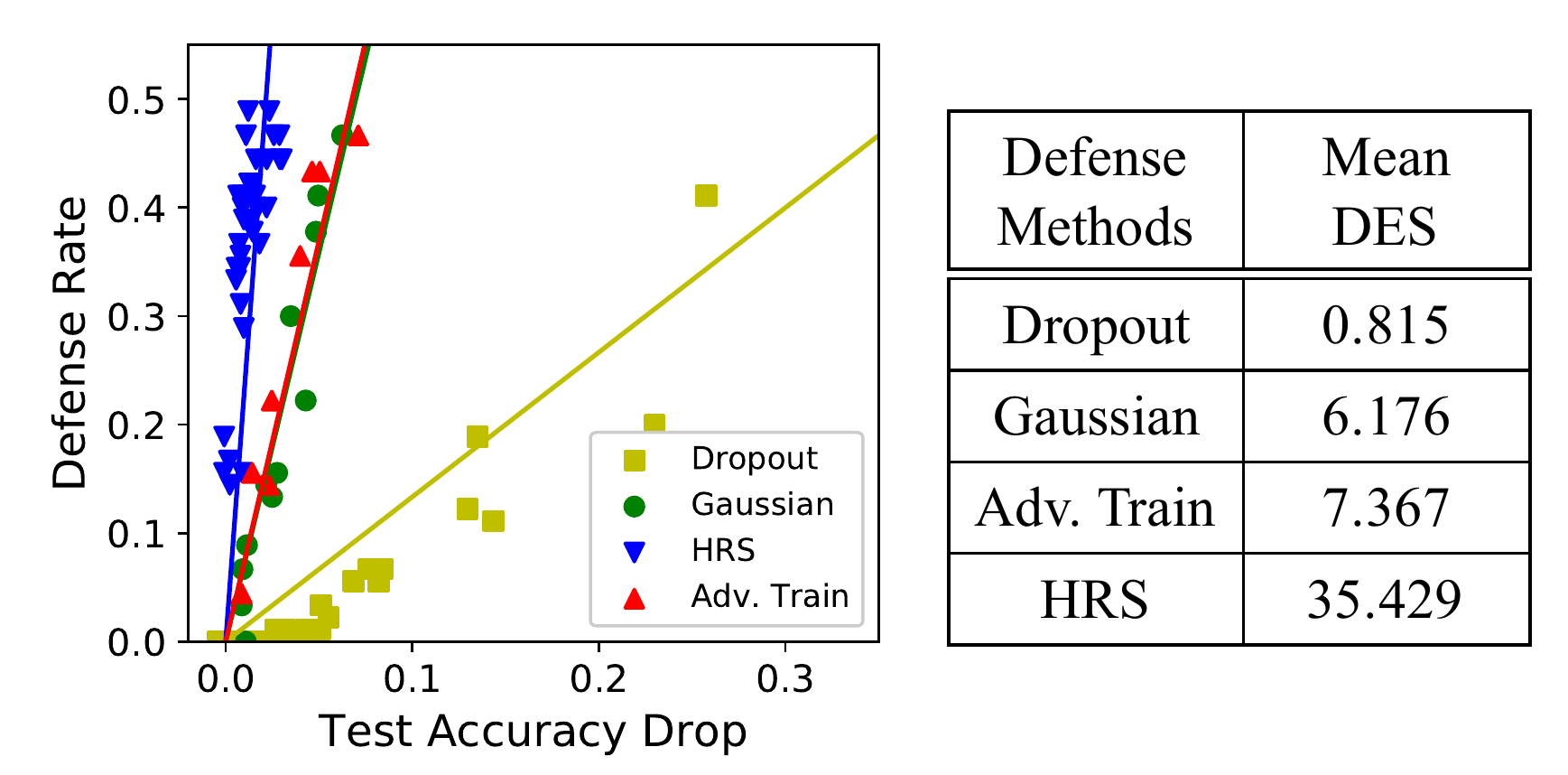}
    \caption{Defense efficiency of different defenses under PGD attack on CIFAR-10. See Appendix \ref{scatter_details} for implementation details. The solid lines are fitted by linear regression.
    }
    \label{fig:DEI_CIFAR}
\end{figure}

We note that although defense effectiveness can be improved by using a stronger defense controlling factor, it is traded by sacrificing the test accuracy on clean examples. We characterize this scenario in Figure \ref{fig:DEI_CIFAR}, where the points of a certain defense method are given by using different strength factors against the same attack. For any tested defense, there is indeed a robustness-accuracy trade-off where stronger defenses are usually associated with more test accuracy drop. For example, on CIFAR-10 and under an $\ell_\infty$ attack strength of 8/255, when adversarial training \cite{madry2017towards} achieves a 56.6\% defense rate, it also causes a 7.11\% drop in test accuracy, which could  be an undesirable trade-off.

Therefore, it is worth noting that even under the same norm-ball bounded adversarial attack threat model, comparing different defense methods is not an easy task as they vary in both defense rate and test accuracy. In order to tackle this difficulty, we propose Defense Efficiency Score (DES) as
\begin{equation}
DES_{D,A}(\theta) = \Delta d / \Delta t
\end{equation}

where $\Delta d$ is the gain in defense rate (percentage of adversarial examples generated by attack $A$ that fails to fool the protected model using defense scheme $D$ with strength factor $\theta$) and $\Delta t$ is the associated test accuracy drop relative to the unprotected base model\footnote{In practice, we use $\Delta d / (\Delta t + \eta)$ where $\eta$ is a small value (we set  $\eta=0.002$) to offset noisy effect (e.g. random training initialization) leading to negative $\Delta t$ when it is close to the origin.}. Intuitively, DES indicates the defense improvement per test accuracy drop.


A fair evaluation of defenses can be conducted by first choosing a desired behavior range (on either defense rate or test accuracy drop), and then compare the statistics (such as mean and variance) of DES values by varying defense strength that fall into the desired range. In Figure \ref{fig:DEI_CIFAR} we show the scatter plot and mean of DES values of HRS (with up to 30 X 30 channels), together with other stochastic defenses and adversarial training that have the same defense rate range. We find that the points (resulting models) of a defense lie roughly on a linear line with a small variance and our proposed HRS defense attains the best mean DES that is more than 3$\times$ higher than the state of the art, suggesting a significantly more effective defense. Details of HRS and DES analysis will be given in Section \ref{sec_HRS} and Section \ref{sec_performance}, respectively.

\begin{figure}[ht]
    \centering
    \includegraphics[width=0.5\textwidth]{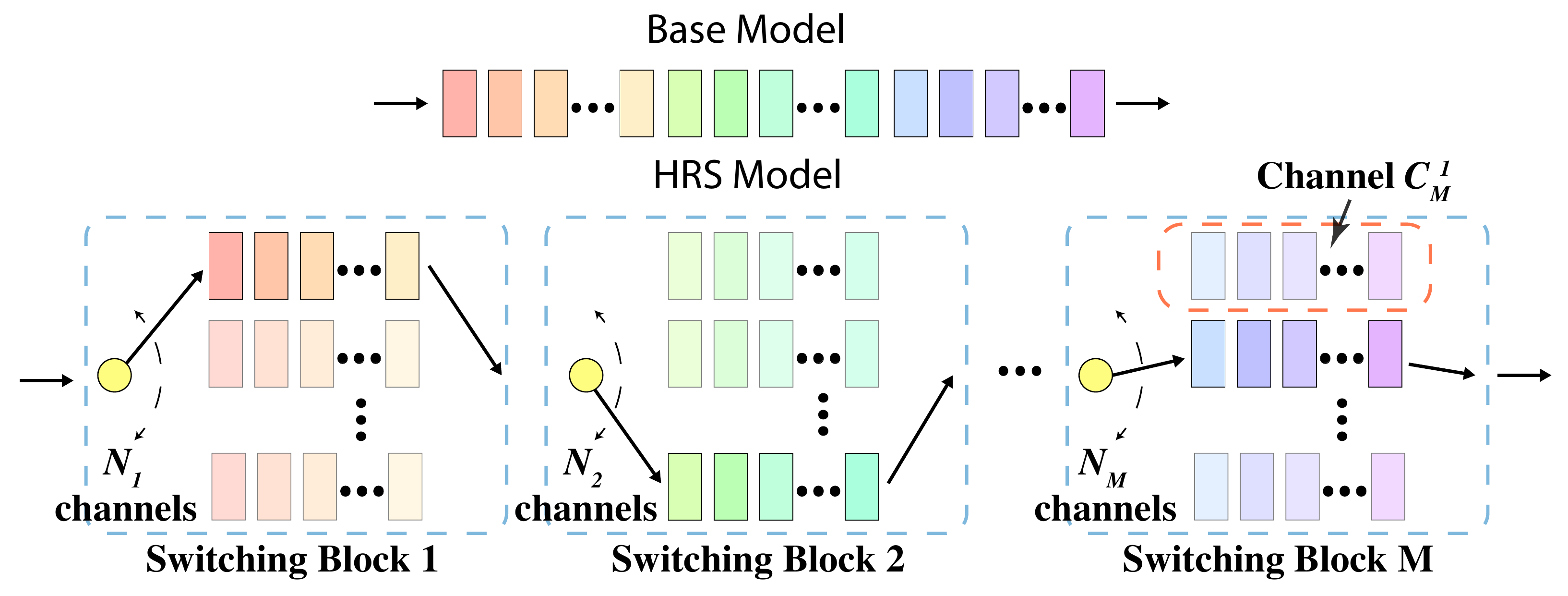}
    \caption{Illustration of HRS-protected model.}
    \label{fig:hrs_show}
\end{figure}

\section{Hierarchical Random Switching (HRS)}
\label{sec_HRS}

\subsection{HRS Protected Model}
HRS divides a base neural network into several \textit{blocks} and replaces each block with a \textit{switching block} which contains a bunch of parallel \textit{channels} with different weights but the same structure as shown in Figure \ref{fig:hrs_show}. HRS features a switcher that randomly assigns the input of the block to one of the parallel channels in the run time. 
We call the selected channel by the switcher an \textit{active} channel, and at any given time all active channels from each block constitute an \textit{active path} which has the same structure as the base model.

Intuitively, a HRS-protected model can assure comparable performance to the base model if each possible path is fully functional while its random switching nature prevents 
the attacker from exploiting the weakness of a fixed model structure. We will introduce a training procedure to ensure full function of each path shortly.  

HRS has two main advantages over current stochastic defenses. First, many defenses introduce randomness by dropping neurons or adding noise, leading to undesirable and even disruptive noisy information for model inference, deteriorating accuracy on legitimate data. 
This explains why these methods have worse trade-offs in terms of DES, as shown in Figure \ref{fig:DEI_CIFAR}.
In contrast, HRS introduces randomness in block switching, where each active path in HRS is trained to have a comparable performance to the base model, which greatly alleviates the issue of significant drop in test accuracy.

Second, HRS is a decentralized randomization scheme. Each variant of HRS has no privilege over others due to random switching. Therefore, it is fundamentally different from Dropout or Gaussian noise where all variations are derived from the base deterministic model, making the base model a centralized surrogate model and potentially leveraged by attackers to bypass these defenses. 
We consider this attack setting as the "fixed-randomness" setting, which is an adaptive white-box attack assuming the attacker knows the base model and is aware of randomness being deployed as defenses. In Section \ref{sec_performance}, we will show our proposed HRS is resilient to such adaptive attack, attaining even better defense performance than standard white-box attacks.



\subsection{Training for HRS}
To facilitate HRS model training and ensure the performance of every path, we propose a bottom-up training approach. The training process start with training for the first switching block by constructing $N_1$ randomly initialized paths. These paths are trained independently to convergence, and after this round of training the weights in the first switching block will be fixed. We then train for the second switching block by constructing $N_2$ paths with randomly initialization except for the first switching block.
During training, the switching scheme of the first block is activated, which forces the following upper blocks to adapt the variation of the first block. 
The training process continues until all switching blocks are trained. We find that by doing so, the training performance is stable, and each channel in a switching block is decentralized. Details of the bottom-up training approach are summarized in Algorithm \ref{algo_HRS} of Appendix \ref{HRS_train}.

\section{Performance Evaluation and Analysis}
\label{sec_performance}

In this section, we run experiments on two datasets, MNIST \cite{lecun1998mnist} and CIFAR-10 \cite{krizhevsky2009learning}, to benchmark our proposed HRS on defending adversarial attacks. The study consists of two parts. In the first part, We test HRS with different channels and three other stochastic network defenses in white-box attack setting (i.e. assume the attacker has full information about the target model including its structure and parameters) and two adaptive attack settings where the attackers attempt to incorporate randomness-aware counter-measures to strengthen their attacks at possibly additional computation cost. Note that for a fair comparison, we need to consider both defense effectiveness and test accuracy. Thus we set defenses to the same accuracy level by tuning their strength controlling factors and report detailed accuracy values in \ref{Test_acc_D1} of Appendix \ref{sup_exp}.

In the second part we provide a comprehensive study on the trade-off between defense performance and drop in test accuracy of each method via the DES introduced in Section \ref{sec_DEI}. We not only compare among different stochastic defense methods but also implememt adversarial training \cite{madry2017towards}, a state-of-the-art deterministic defense method\footnote{We only compare adversarial training  on CIFAR-10, as on MNIST it does not suffer from large test accuracy drop.}.


\subsection{Experiment Settings}
\subsubsection{Base Network Models}
For a fair comparison all defenses have to be applied on the same unprotected model (base model). We use two convolutional neural network (CNN) architectures as our base models for MNIST and CIFAR-10 datasets, respectively, as they were standard settings considered in previous works such as \cite{carlini2017towards,papernot2016distillation,chen2017ead}. Details about these base models are summarized in Appendix \ref{Model_Architecture}.



\subsubsection{Stochastic Network Defense Schemes}
Below summarizes the implemented defenses, and Table \ref{Table: Test_acc} in Appendix \ref{sup_results} shows their resulting test accuracy. 


\begin{itemize}[leftmargin=*]
\item \textbf{SAP} \cite{s.2018stochastic} : Stochastic activation pruning (SAP) scheme is implemented on the base model between the first and second fully-connected layers.

\item \textbf{Defensive Dropout} \cite{wang2018defensive}: dropout is used between the first and second fully-connected layers with three different dropout rates, $0.1$, $0.3$ and $0.7$ ($0.7$ is omitted on MNIST as it severely degrades test accuracy). 

\item \textbf{Gaussian Noise}: Following the setting in \cite{liu2017towards}, we add Gaussian noise to the input of each convolutional layer in both training and testing phases. We defer its parameter setting and discussion to Appendix \ref{sec_def_implement}.
	
\item \textbf{HRS} (proposed): We divide the base model structure into two switching blocks between the first and second fully-connected layers. We implement this 2-block HRS-protected model with $10 \times 10$, $20 \times 20$ and $30 \times 30$ channels. Note that at any given time, the active path of HRS has the same structure as the base model.
	
\end{itemize}

\subsubsection{Attack Settings}

We consider the standard white-box attack setting and two adaptive white-box attacks settings (expectation over transformation (EOT)\cite{athalye2017synthesizing,athalye2018obfuscated} and fixed randomness) for stochastic defenses. 
The purpose of using EOT and fixed randomness attacks is to show the defense effectiveness is not a consequence of obfuscated gradients.
In each setting, four adversarial attack methods are implemented: 
 FGSM, CW, PGD and CW-PGD. Their implementation details are summarized in Appendix \ref{sup_attacks}. 
\begin{itemize}[leftmargin=*]
    \item \textbf{White-box Attack:} The adversary uses the stochastic model directly to generate adversarial examples.
    \item \textbf{Expectation Over Transformation (EOT):}
 When computing input gradient, the adversary samples input gradient for $n$ times and use the mean of gradients to update the perturbed example. We set $n=10$ in our experiments as we observe no significant gain when using $n>10$. Details about the pilot research on $n$ are given in Appendix \ref{scatter_details}.
    \item \textbf{Fixed Randomness:}  Generating adversarial examples using a fixed model by disabling any randomness scheme. For  SAP, defensive dropout and Gaussian noise, it is done by removing their randomness generation modules (e.g. dropout layers). For HRS, it is done by fixing an active path.
\end{itemize}

\begin{figure}[t]
    \centering
    \includegraphics[width=0.5\textwidth]{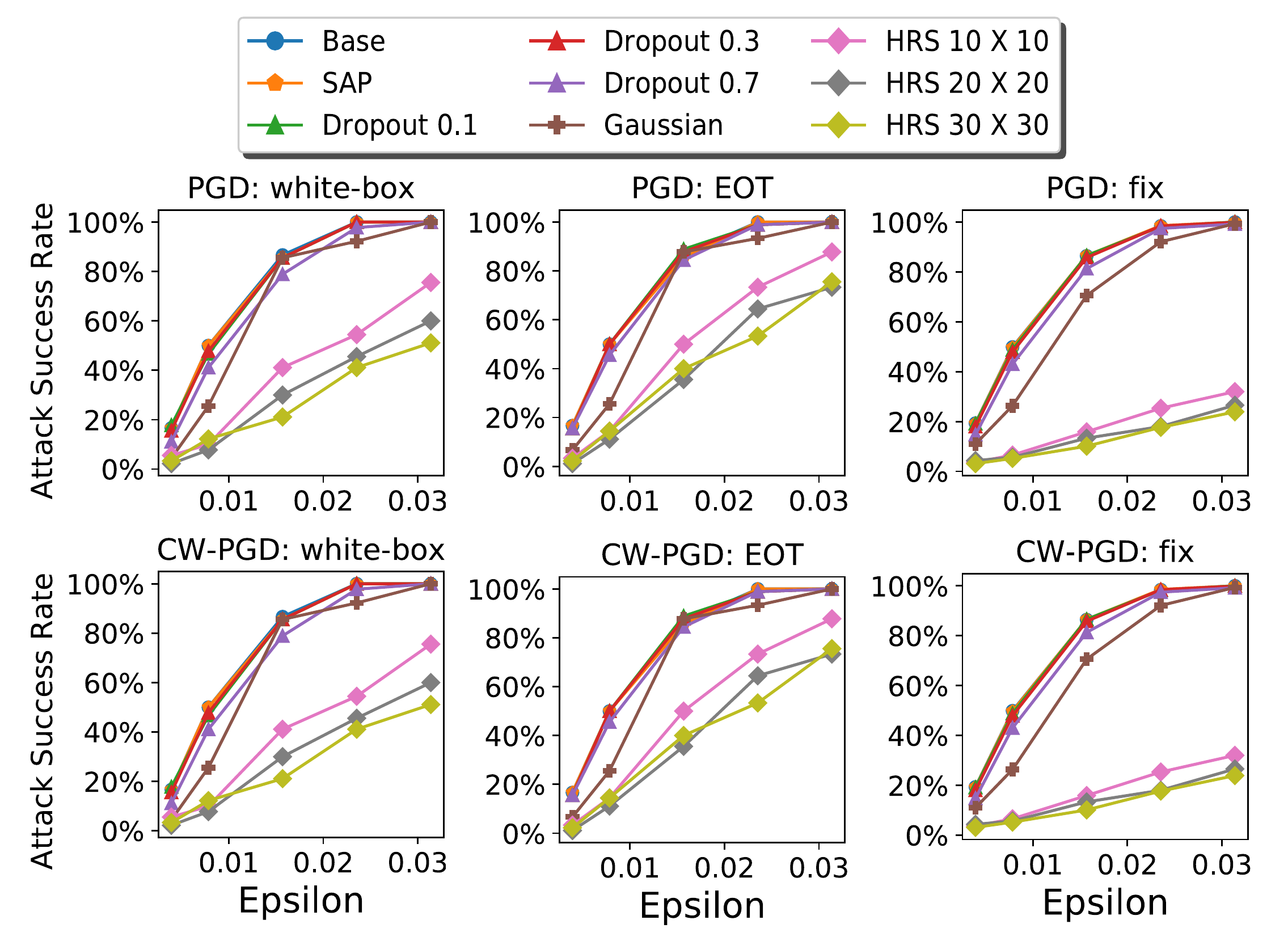}
    \caption{Attack success rate on CIFAR-10 using (a) PGD, (b) PGD + EOT, (c) PGD + fixed randomness (d) CW-PGD, (e) CW-PGD + EOT, and (f) CW-PGD + fixed randomness.}
    \label{fig:attacks(pgd&cwpgd)}
\end{figure}

\subsection{White-box and Adaptive Attack Analysis}
\label{6.2_defense}
Due to space limitation, we compare the attack success rate (ASR) of different defenses on CIFAR-10 dataset against PGD and CW-PGD attacks with different strengths (the $\ell_\infty$ constraint) and under three attack settings in Figure \ref{fig:attacks(pgd&cwpgd)}. We defer the experimental results of FGSM and CW and all attacks on MNIST dataset to Appendix \ref{sup_attacks}.



We summarize our findings from experiments as follows:
\begin{enumerate}


\item HRS achieves superior defense performance under all attack settings. The advantage of HRS over other defenses becomes more apparent under stronger attacks such as PGD and CW-PGD, where SAP, defensive dropout and Gaussian noise provide little defense given the same level of test accuracy drop. For example, even with a reasonably large $\ell_\infty$ perturbation constraint $8/255$, on CIFAR-10 PGD and CW-PGD attacks only attain 51.1\% and 54.5\% ASRs on HRS with only at most 0.48\% drop in test accuracy, respectively, while all other stochastic defenses are completely broken (100\% ASR), and adversarial training  with the same defense rate has  7\% more test accuracy drop.

\item The adaptive attacks using EOT can marginally improve ASR on all defenses when compared with the standard white-box attacks. But it is not as efficient as fixed-randomness attack or white-box attack as it requires $n$ times gradient computations.

\item We observe that the fixed randomness adaptive attack leads to distinct consequences to different defense methods. For SAP, dropout and Gaussian noise, it has a similar effect as using EOT but without requiring multiple input gradient samples. However, for HRS it actually has a worse performance than standard white-box attacks. This phenomenon can be explained by the decentralized randomness property of HRS as discussed in Section \ref{sec_HRS}.

\item We observe that HRS with more channels are stronger in defense. Note that HRS introduces little computation overhead than the base model and it has large DEI score. In practice, further reduction of ASR with HRS can be achieved by simply allowing more channels or blocks.




\end{enumerate}

\subsection{Defense Efficiency Analysis}

To characterize the robustness-accuracy trade-offs of different defenses, we compare the DES (see Section \ref{sec_DEI}) of different stochastic defense methods in Table \ref{Table: DEI}. Our HRS attains the highest DES on CIFAR-10 for all attacks and on MNIST for most attacks. In particular, HRS outperforms other defenses by a large margin (at least by $3\times$) on CIFAR-10.



\begin{figure}[t]
\centering
\includegraphics[width=0.48\textwidth]{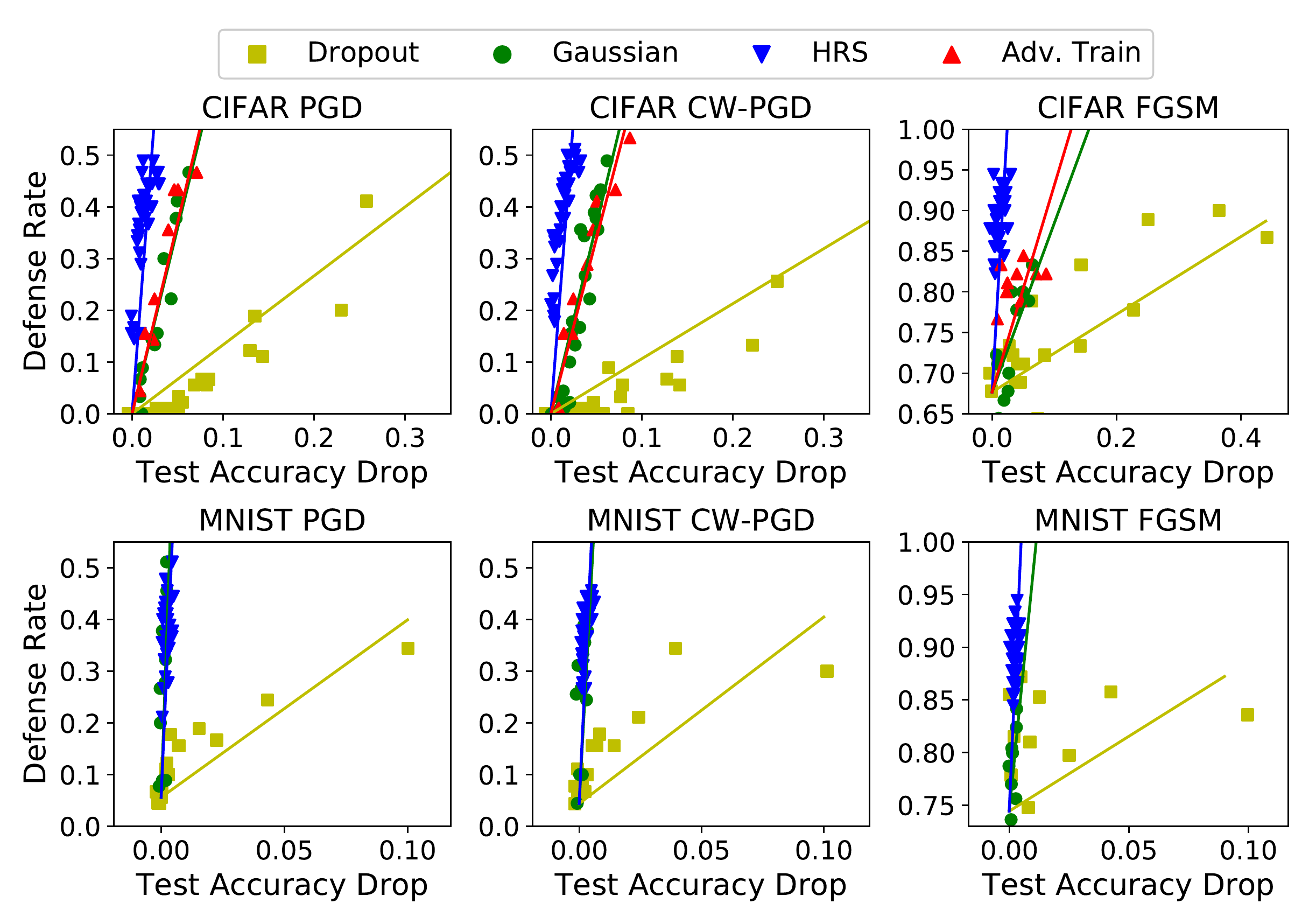}
\caption{Scatter plots of different defenses. Attacks on CIFAR-10 and MNIST are using 8/255 and 64/255 $\epsilon$ bounds, respectively.}
\label{fig:cifar_scatter}
\end{figure}

\begin{table}[t]
\begin{center}
\label{Table: DEI}
\scalebox{0.9}{
\begin{tabular}{|l|c|c|c|c|}
\hline
 Dataset & Defense & FGSM  & PGD  & CW-PGD \\
\hline\hline
\multirow{3}{1.5cm}{MNIST} & Dropout & 11.99 & 17.94 & 19.55 \\
& Gaussian & 14.90 & 71.78 & \textbf{82.76} \\
& HRS & \textbf{36.41} &\textbf{76.64} & 74.90 \\
\hline
\multirow{3}{1.5cm}{CIFAR-10} & Dropout & 1.09 & 0.81 & 0.76 \\
& Gaussian & 2.73 & 6.17 & 5.47 \\
& Adv. Train & 5.05 & 7.37 & 6.57 \\
& HRS & \textbf{32.23} & \textbf{35.43} & \textbf{35.55} \\
\hline
\end{tabular}}
\end{center}
\caption{Mean DES of different defense methods}
\end{table}

\section{First Defense against Adversarial Reprogramming}
\label{sec_adv_reprog}

\begin{figure}
    \centering
    \includegraphics[width=0.45\textwidth]{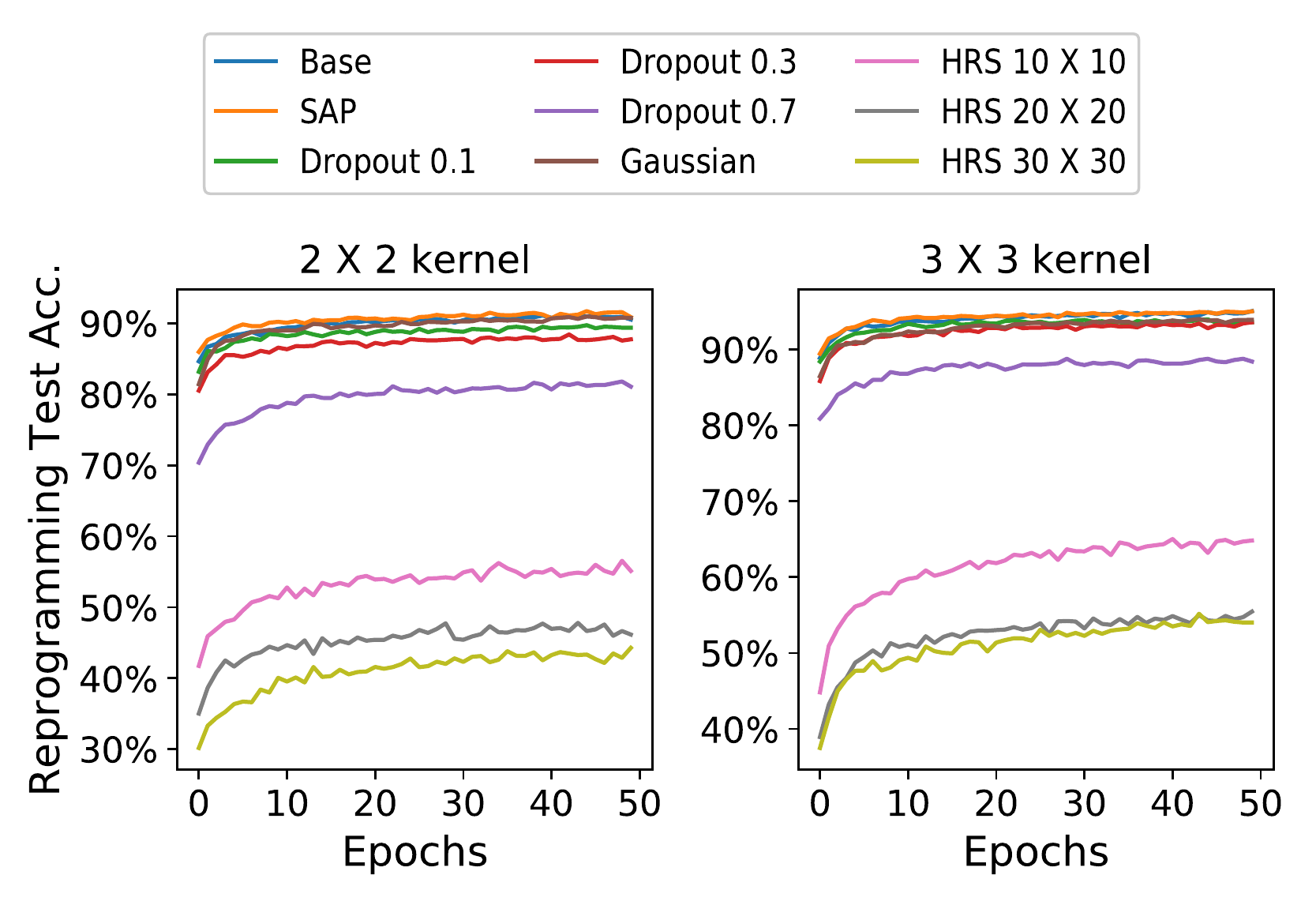}
    \caption{Adversarial reprogramming test accuracy during training a locally connected layer with different kernels as input transform. }
    \label{fig:ADV_Reprogramming}
\end{figure}

In addition to defending adversarial misclassification attacks, here we demonstrate the effectiveness of HRS against adversarial reprogramming  \cite{elsayed2018adversarial}. We use the same base network on CIFAR-10  in Section \ref{sec_performance} as the target model to be reprogrammed to classify MNIST images. We use a locally connected layer to perform the input transformation with different kernel sizes and use an identical mapping as the output transformation. 
The unprotected classifier can easily be reprogrammed to achieve up to 90.53\% and 95.07\% test accuracy using kernel sizes $2\times2$ and $3\times3$, respectively, on classifying MNIST images after several epochs of training for the input transformation.

We compare the defenses against adversarial reprogramming using the same set of defense methods in Section \ref{sec_performance} and show the reprogramming test accuracy during 50 epochs of training in Figure \ref{fig:ADV_Reprogramming}. We observe that HRS-protected models can significantly reduce the adversarial reprogramming test accuracy whereas all other defenses have less defense effect. 




\section{Conclusion}

To fairly characterize the robustness-accuracy trade-offs of different defenses, in this paper we propose a novel and comprehensive metric called defense efficiency score (DES). In addition, towards achieving a better trade-off, we propose hierarchical random switching (HRS) for defense, which can be easily compatible with typical network training procedures. Experimental results show that HRS has superior defense performance against standard and adaptive adversarial misclassification attacks while attaining significantly higher DES than current stochastic network defenses. HRS is also the first effective defense against adversarial reprogramming.

\clearpage
\bibliographystyle{named}
\bibliography{ref}

\clearpage
\appendix

\setcounter{figure}{0}
\setcounter{table}{0}
\makeatletter
\renewcommand{\thefigure}{A\arabic{figure}}
\renewcommand{\thetable}{A\arabic{table}}

\section{Motivation of Using Randomness}
\label{sup_motivation}
\subsection{Motivation 1}
\label{sup_motivation1}
\begin{figure}[htbp]
    \centering
    
    \includegraphics[width=0.35\textwidth]{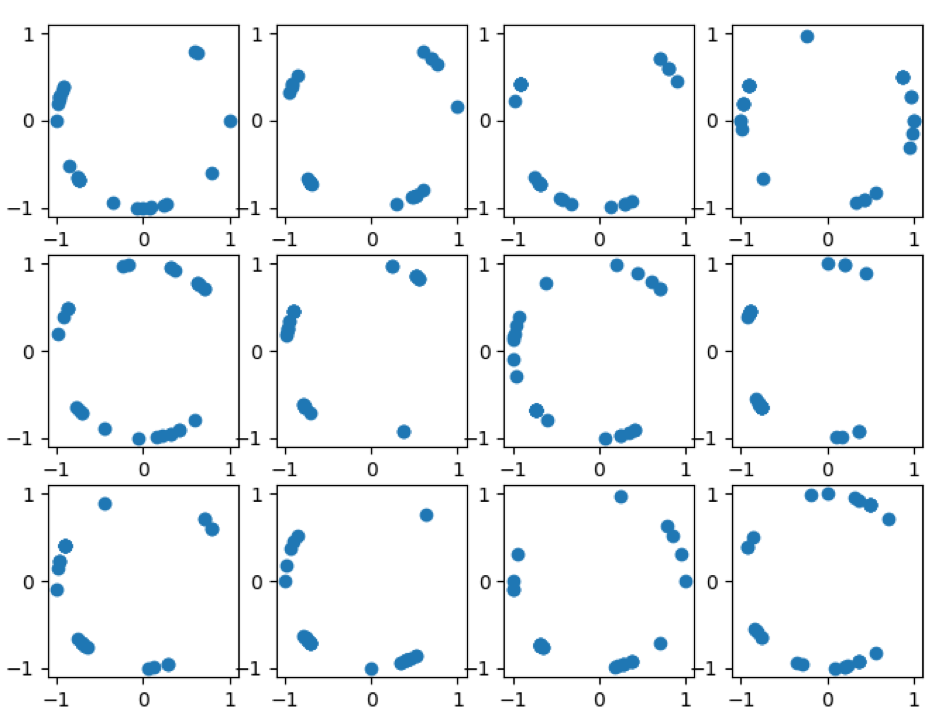}
    \caption{Distribution of adversarial examples for a group of model individuals of the same architecture. Each grid shows the relative positions of adversarial examples to the clean image (the origin) on two 
    randomly selected input dimensions.
    Here, the distances of adversarial examples on the given x-y plane are normalized. Adversarial examples are generated by CW-PGD attack.}
    \label{motivation3}
\end{figure}

We find that when training models with the same network architecture but with different weight initialization, each model has its own vulnerable spots.
Figure \ref{motivation3} shows the adversarial examples found by the same attack for each model. These models all have the same architecture, training process and performance in terms of test accuracy and defense capability. However, it is surprising that the distribution of adversarial examples for each model is not concentrated but quite scattered. This indicates that each model has its own robustness characteristics even though they are identical on the macro scope.


\subsection{Motivation 2}
\label{sup_motivation2}
\begin{figure}[hb]
    \centering
    \begin{minipage}[b]{0.36\textwidth}
\centering  
\includegraphics[width=1\textwidth]{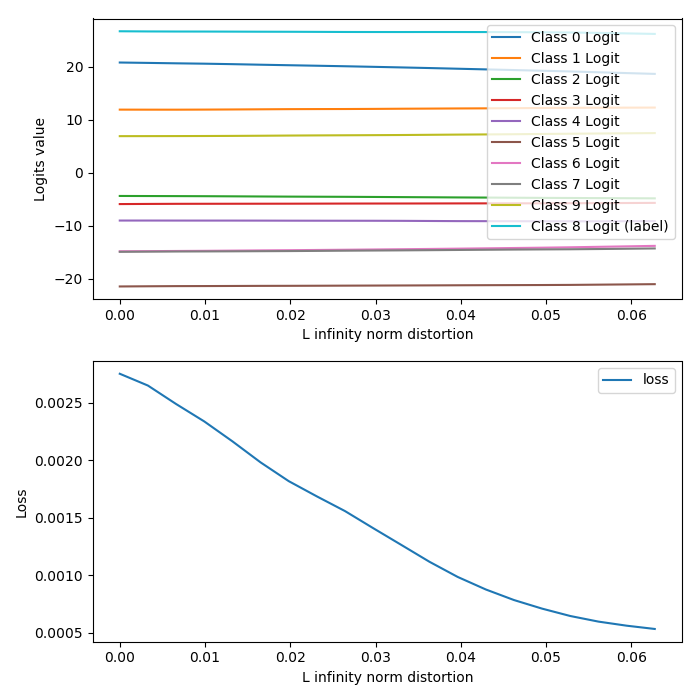} \\
\end{minipage}
    \begin{minipage}[b]{0.36\textwidth}
\centering  
\includegraphics[width=1\textwidth]{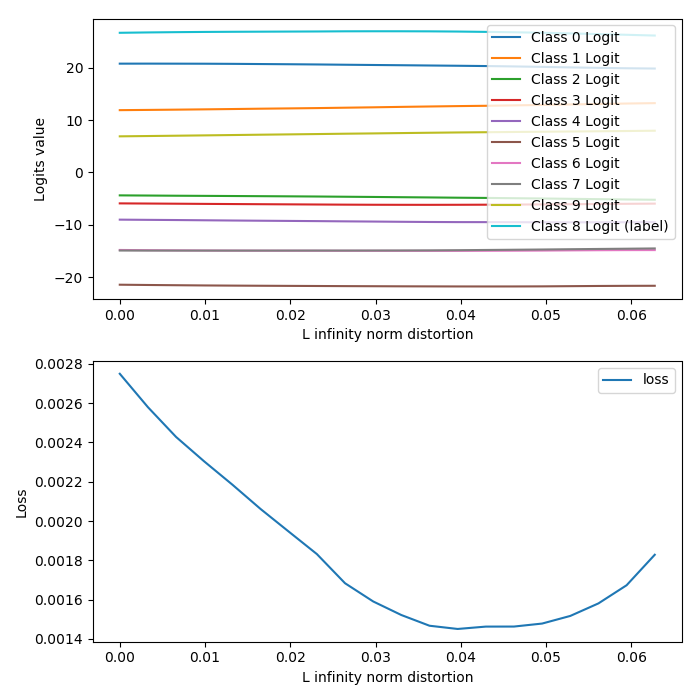} \\
\end{minipage}
    \vspace{-2mm}
    \caption{Examples of logits and loss changes in random directions.}
    \label{Figure:motivation}
\end{figure}

\begin{figure}[ht]
    \centering
    \begin{minipage}[b]{0.36\textwidth}
\centering  
\includegraphics[width=1\textwidth]{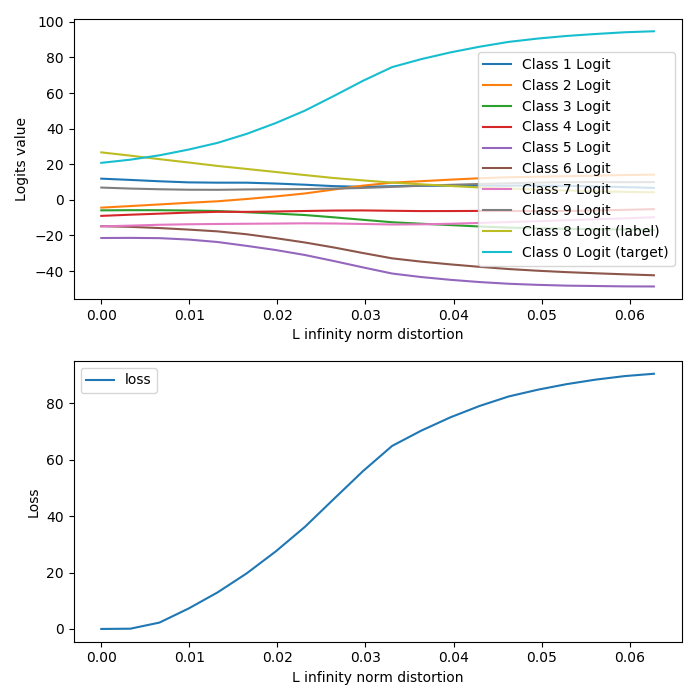} \\
\end{minipage}
    \begin{minipage}[b]{0.36\textwidth}
\centering  
\includegraphics[width=1\textwidth]{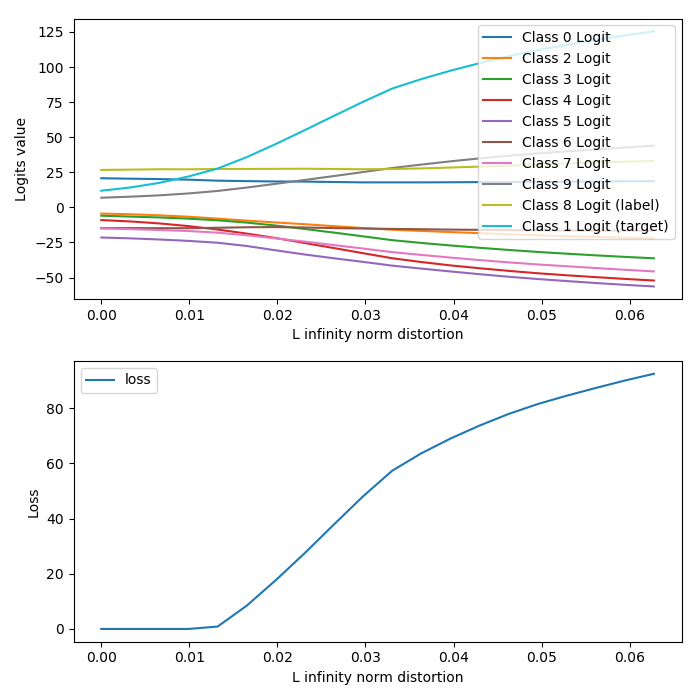} \\
\end{minipage}
    \caption{Examples of logits and loss changes in (targeted) adversarial directions. Adversarial directions are found by performing CW-PGD attack on the clean image (origin).}
    \label{Figure:motivation2}
\end{figure}

We find that following a randomly selected direction, the possibility  of finding a successful adversarial example is very low. 
Figure \ref{Figure:motivation} shows the changes of the logit (pre-softmax) layer representations and the training losses. One can observe that their changes are quite subtle and thus the model is actually quite robust to perturbations guided by random directions. In contrast, when comparing to the changes of an adversarial direction made by interpolating between a pair of original example and adversarial example found by performing targeted CW-PGD attack, the logit of the target class has a notable increase, shown in Figure \ref{Figure:motivation2}. This indicates that although DNNs can be robustness to random perturbations, it still lacks worst-case robustness and thus becomes vulnerable to adversarial attacks.

\subsection{Input Gradient Deviation v.s. Mean Defense Rate}
\label{sup_deviation}
\begin{figure}[!hb]
    \centering
    \includegraphics[width=0.4\textwidth]{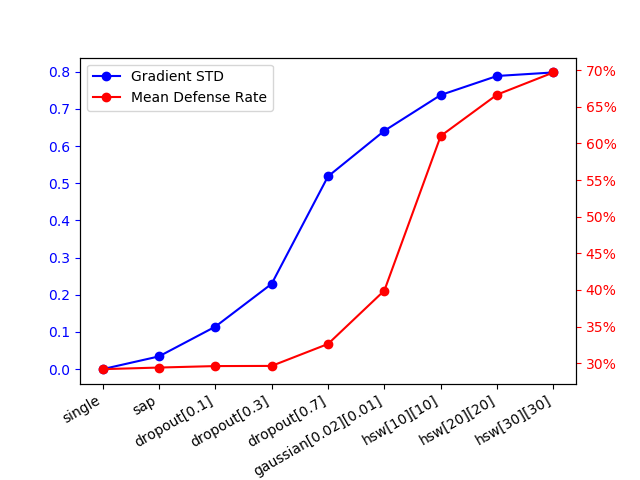}
    \caption{Input gradient variance is highly correlated with defense effectiveness for stochastic defenses.}
    \label{gradient_dev_fig}
    \vspace{-5mm}
\end{figure}




\clearpage

\section{Training for HRS}
\label{HRS_train}
\vspace{-2mm}
\begin{algorithm}[htbp]
\caption{Training HRS-protected Model}
\label{algo_HRS}
\begin{algorithmic}[1]
\REQUIRE ~~\\
HRS model architecture with $M$ switching blocks. $N^i$ denotes the number channels in block $i$ and $c^j_i$ denotes the $j$'s channel in block $i$.
\FOR{block $i \in [1,M]$}
\FOR{channel $j \in [1,N_i]$}
\STATE Construct a HRS Model with $M$ blocks:
\FOR{block $k \in [1,M]$}
\IF{$k < i$}
\STATE
Construct $N_i$ channels with trained channels $\{c^l_k \mid l \in [1,N_k] \}$;
\STATE
\textbf{Freeze} channels in block $k$;
\ELSE
\STATE
Construct $1$ channel for block $k$ with randomly initialized weights;
\STATE
Set all channels in block $k$ to be \textbf{trainable};
\ENDIF
\STATE
Train all trainable channels to convergence;
\STATE
Save trained channel $c^j_i$;
\ENDFOR
\ENDFOR
\ENDFOR
\RETURN
A HRS Model with trained channels $\{c^j_i \mid i \in [1, M], j \in [1,N_i] \}$
\end{algorithmic}
\end{algorithm}
\vspace{-2mm}

\clearpage
\section{Experiment Details}
\label{sup_exp}
\subsection{Attack Details}
\label{sup_attacks}

All attacks are implemented in a white-box, targeted attack setting for a fair comparison. For reproducibility of the experiments, we summarize the hyper-parameters we used for each attack.

\begin{itemize}
    \item \textbf{FGSM:} No hyper-parameter needs to specify. Different attack strengths are given by varying the step size $\epsilon$.
    \item \textbf{CW:} We run CW attack with $L_2$ distortion metric. We run gradient descent for 100 iterations with step size of 0.1 and use 10 rounds binary search finding the optimal weight factor $c$. Different attack strengths are given by varying the confidence factor $\kappa$.
    
    \item \textbf{PGD}: We run gradient descent for 100 iterations with step size of 0.1. Different attack strengths are given by varying maximum allowed $L_\infty$ perturbation $\epsilon$.
    \item \textbf{CW-PGD}: The same setting as PGD.
\end{itemize}


\subsection{Defense Implementation of Gaussian Noise}
\label{sec_def_implement}
	On MNIST, we use the recommended standard deviations which are 0.2 for the "init-noise" (noise before the input layer) and 0.1 for the "inner-noise" (noise before other conv layers). However, on CIFAR-10 we found this setting decreases test accuracy significantly (reducing to ~60\%), thus we use $10\times$ smaller deviations (0.02 and 0.01 respectively). We also found that using Gaussian noise solely is not sufficient to prevent the model from over-fitting. As a solution we also use dropout during training in order to prevent over-fitting. 

\subsection{Pilot Research on EOT}
\begin{figure}[htbp]
    \vspace{-3mm}
    \centering
    \includegraphics[width=0.45\textwidth]{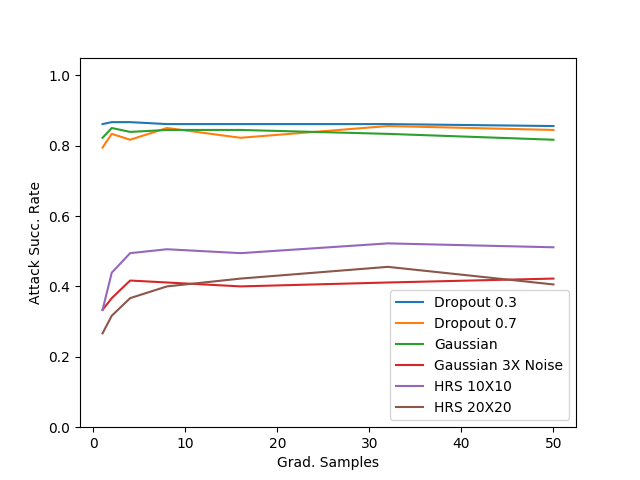}
    \vspace{-2mm}
    \caption{Pilot Research on EOT.}
    \label{fig:n_search}
    \vspace{-2mm}
\end{figure}

We run a pilot test to determine the value of $n$ for EOT attacks. We find using $n=10$ is enough as the benefits of using a larger $n$ saturates when $n>10$. An example of using different $n$ values for EOT is given in Figure \ref{fig:n_search}.
Here we also plot Gaussian defense with a $3\times$ larger noise deviation which is not in Section \ref{6.2_defense} as it drops the test accuracy to 76.83\%. The purpose is to show that the defenses that seem to be less sensitive to EOT (as shown by the lines on the top of Figure \ref{fig:n_search}) do not indicate they are truly resistant to EOT.

\subsection{Model Architecture}
\label{Model_Architecture}
 \begin{table}[htbp]
\caption{Base model architectures for MNIST and CIFAR-10 Datasets.}
\label{table_model_architecture}
\vspace{-5mm}
\begin{center}
\scalebox{0.79}{
\begin{tabular}{|c|c|c|}
\hline
     & Model for MNIST  &  Model for CIFAR-10\\
    \hline
     Conv layer & 32 filters with size (3,3) & 64 filters with size (3,3) \\
     Conv layer & 32 filters with size (3,3) & 64 filters with size (3,3) \\
     Pooling layer & pool size (2,2) & pool size (2,2) \\
     Conv layer & 64 filters with size (3,3) & 128 filters with size (3,3) \\
     Conv layer & 64 filters with size (3,3) & 128 filters with size (3,3) \\
     Pooling layer & pool size (2,2) & pool size (2,2) \\
     Fully connected & 200 units & 256 units \\
     Fully connected & 200 units & 256 units \\
     Output layer & 10 units & 10 units \\
     \hline
\end{tabular}}
\vspace{-5mm}
\end{center}
\end{table}

\subsection{Implementation Details on Study of Defense Efficiency}
\label{scatter_details}
Spots of HRS are due to different number of block channels ranging from $5\times5$ to $30\times30$. Spots of dropout are due to different training and testing dropout rate ranging from 0.1 to 0.9. Spots of Gaussian noise are due to different initial and inner Gaussian noise deviations ranging from (0.01, 0.005) to (0.11, 0.055) on CIFAR-10 and from (0.1, 0.05) to (0.325, 0.1625) respectively. Spots of Adversarial Training are due to different $\epsilon$ bounds from 0.5/255 to 4.5/255 of adversarial examples used in training.

\subsection{Test Accuracy}
\label{Test_acc_D1}
\begin{table}[hb]
\begin{center}
\caption{Test accuracy of different defense methods.}
\label{Table: Test_acc}
\scalebox{0.9}{
\begin{tabular}{|l|c|c|c|c|}
\hline
Model & MNIST & Dev.(e-4)  & CIFAR  & Dev.(e-4) \\
\hline\hline
Base & 99.04\% & / &79.17 \% & / \\
SAP & 99.02\% & 1.47 &79.16 \% & 2.81 \\
Dropout 0.1 & 98.98\% & 3.45 & 79.08\% & 7.99\\
Dropout 0.3 & 98.68\% & 6.06 & 78.65 \% & 16.67\\
Dropout 0.7 & /  & / & 76.02 \%& 24.52 \\
Gaussian & 99.02\% & 5.82 & 78.04 \% & 7.25\\
HRS 10*10 & 98.95\% & 5.33 & 78.93\% & 26.81 \\
HRS 20*20 & 98.91\% & 6.02 & 78.76\% & 20.32\\
HRS 30*30 & 98.85\% & 8.31 & 78.69\% & 23.25\\
\hline
\end{tabular}}
\end{center}
\vspace{-4mm}
\end{table}

\clearpage
\section{Experimental Results}
\label{sup_results}

\subsection{MNIST}
\begin{figure}[!hb]
    \centering
    \vspace{-4mm}
    \subfigure{
\centering 
\begin{minipage}[b]{0.45\textwidth}
\centering  
\includegraphics[width=1\textwidth]{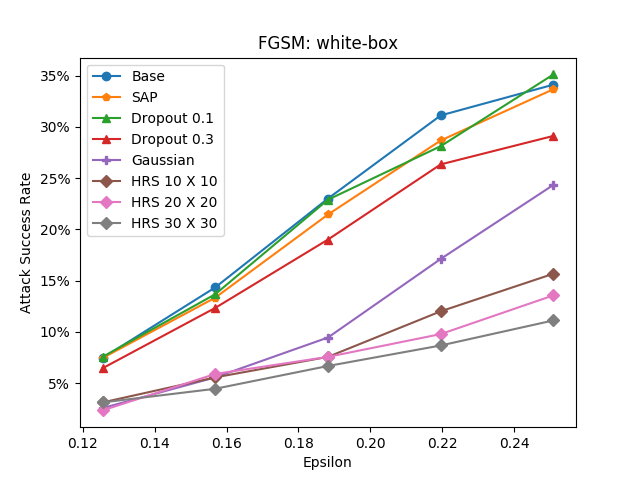} \\
\end{minipage}
}
    \vspace{-4mm}
    \subfigure{
\centering 
\begin{minipage}[b]{0.45\textwidth}
\centering  
\includegraphics[width=1\textwidth]{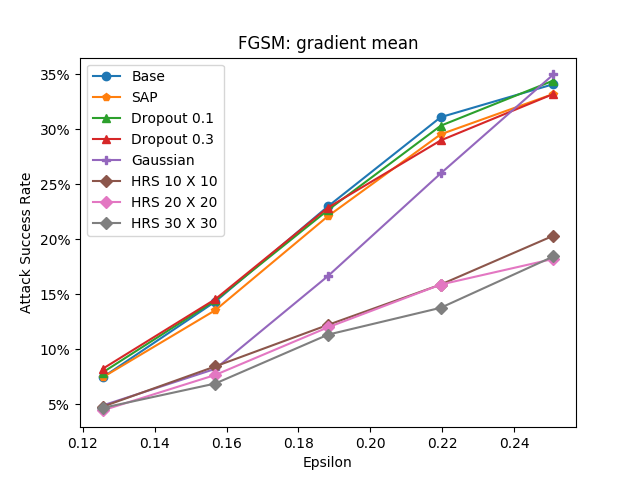}
\end{minipage}
}
    \vspace{-4mm}
    \subfigure{
\centering 
\begin{minipage}[b]{0.45\textwidth}
\centering  
\includegraphics[width=1\textwidth]{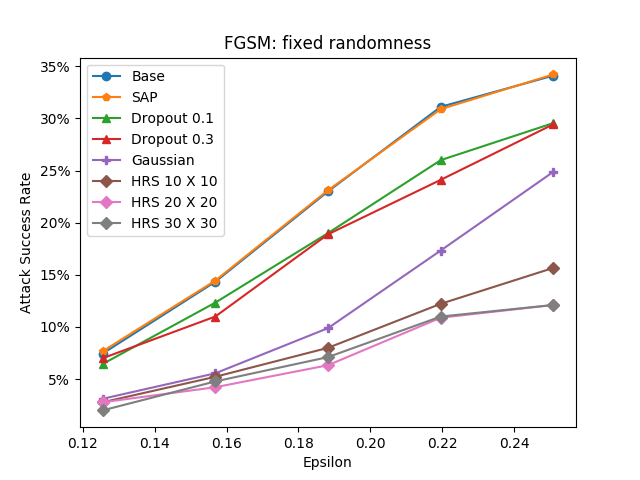}\\
\end{minipage}
}
\caption{Attack success rate of FGSM on different defenses.}
    \label{Figure:MNIST_FGSM}
    \vspace{-2mm}
\end{figure}

\begin{figure}[ht]
    \centering
    \vspace{-19mm}
    \subfigure{
\centering 
\begin{minipage}[b]{0.45\textwidth}
\centering  
\includegraphics[width=1\textwidth]{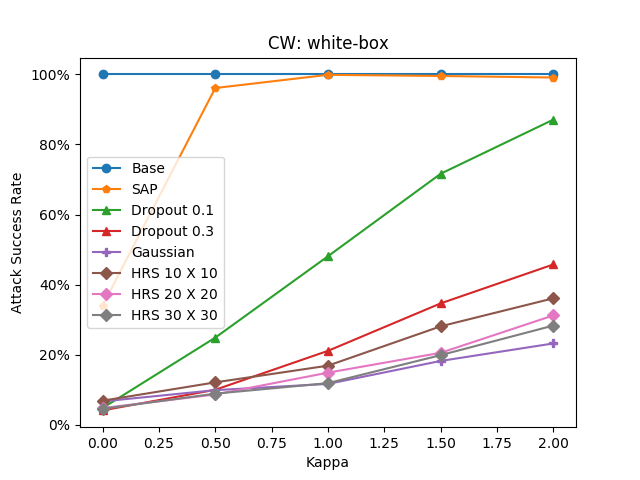} \\
\end{minipage}
}
   \vspace{-4mm}
    \subfigure{
\centering 
\begin{minipage}[b]{0.45\textwidth}
\centering  
\includegraphics[width=1\textwidth]{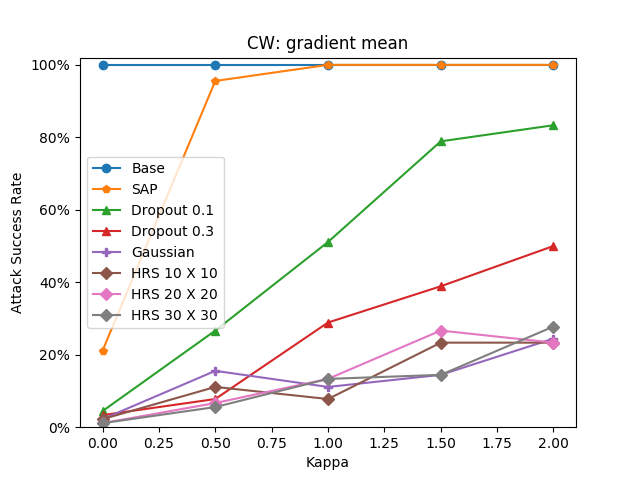}
\end{minipage}
}   
    \vspace{-4mm}
    \subfigure{
\centering 
\begin{minipage}[b]{0.45\textwidth}
\centering  
\includegraphics[width=1\textwidth]{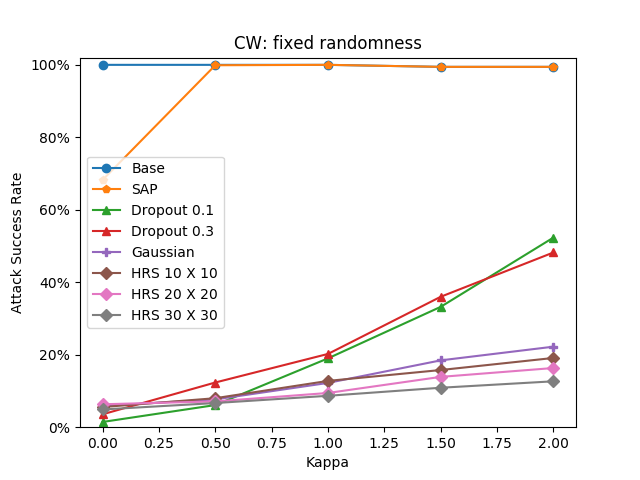}\\
\end{minipage}
}
\caption{Attack success rate of CW on different defenses.}
    \label{Figure:MNIST_CW}
\end{figure}

\begin{figure}[ht]
    \centering
    \subfigure{
\centering 
\begin{minipage}[b]{0.45\textwidth}
\centering  
\includegraphics[width=1\textwidth]{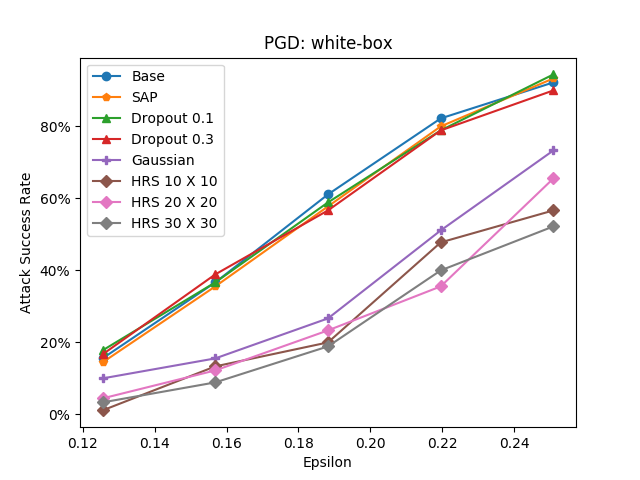} \\
\end{minipage}
}
    \subfigure{
\centering 
\begin{minipage}[b]{0.45\textwidth}
\centering  
\includegraphics[width=1\textwidth]{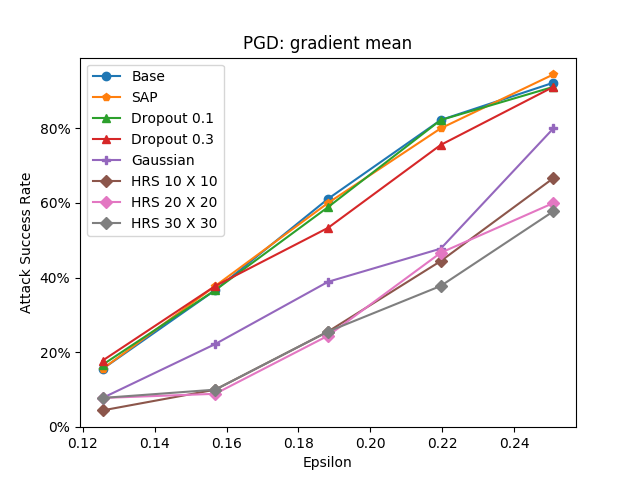}
\end{minipage}
}
    \subfigure{
\centering 
\begin{minipage}[b]{0.45\textwidth}
\centering  
\includegraphics[width=1\textwidth]{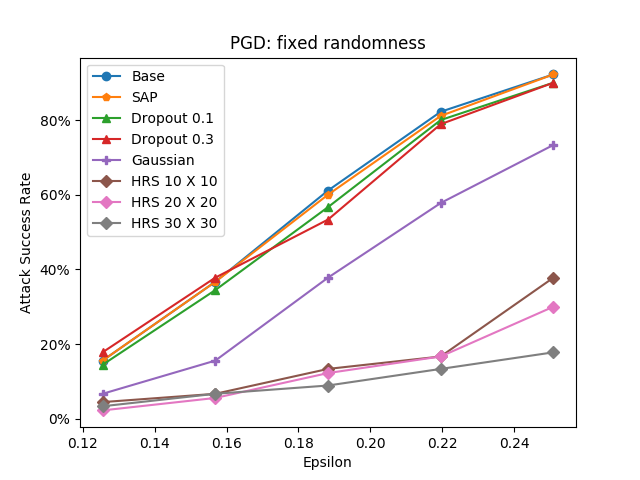}\\
\end{minipage}
}
\caption{Attack success rate of PGD on different defenses.}
    \label{Figure:MNIST_PGD}
\end{figure}

\begin{figure}[ht]
    \centering
    \subfigure{
\centering 
\begin{minipage}[b]{0.45\textwidth}
\centering  
\includegraphics[width=1\textwidth]{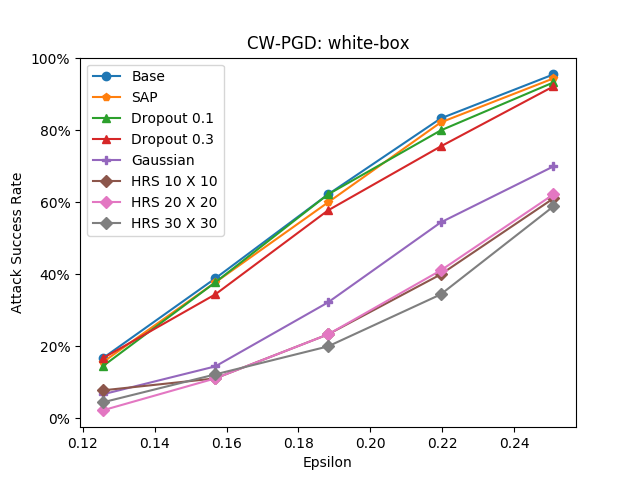} \\
\end{minipage}
}
    \subfigure{
\centering 
\begin{minipage}[b]{0.45\textwidth}
\centering  
\includegraphics[width=1\textwidth]{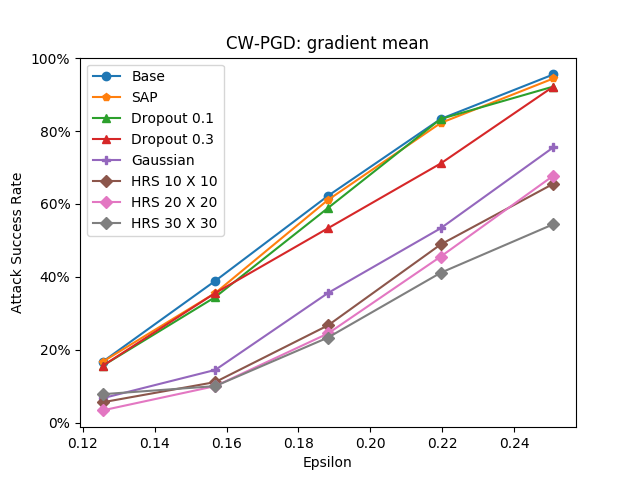}
\end{minipage}
}
    \subfigure{
\centering 
\begin{minipage}[b]{0.45\textwidth}
\centering  
\includegraphics[width=1\textwidth]{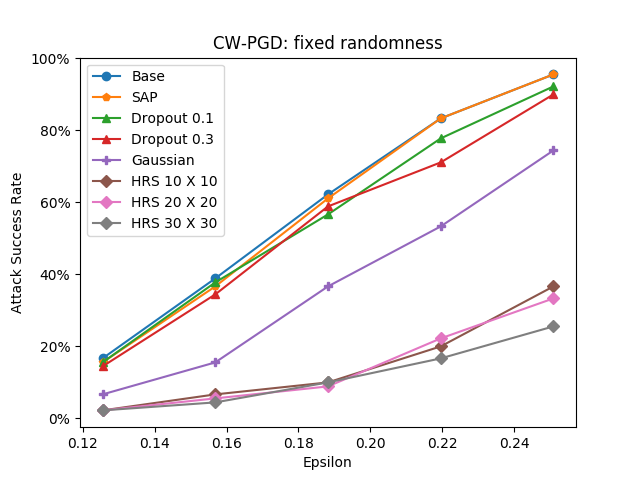}\\
\end{minipage}
}
\caption{Attack success rate of CW-PGD on different defenses.}
    \label{Figure:MNIST_CW2}
\end{figure}

\clearpage
\subsection{CIFAR-10}
\begin{figure}[!hb]
    \centering
    \subfigure{
\centering 
\begin{minipage}[b]{0.45\textwidth}
\centering  
\includegraphics[width=1\textwidth]{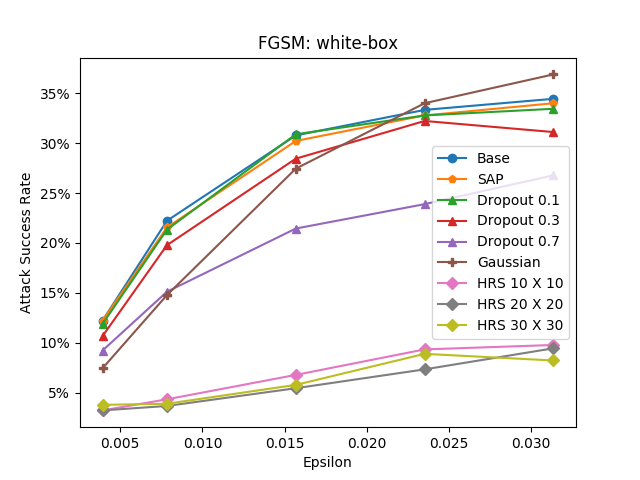} \\
\end{minipage}
}
    \subfigure{
\centering 
\begin{minipage}[b]{0.45\textwidth}
\centering  
\includegraphics[width=1\textwidth]{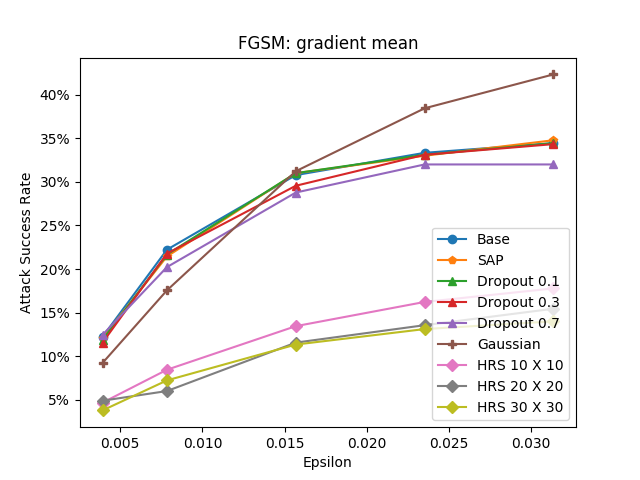}
\end{minipage}
}
    \subfigure{
\centering 
\begin{minipage}[b]{0.45\textwidth}
\centering  
\includegraphics[width=1\textwidth]{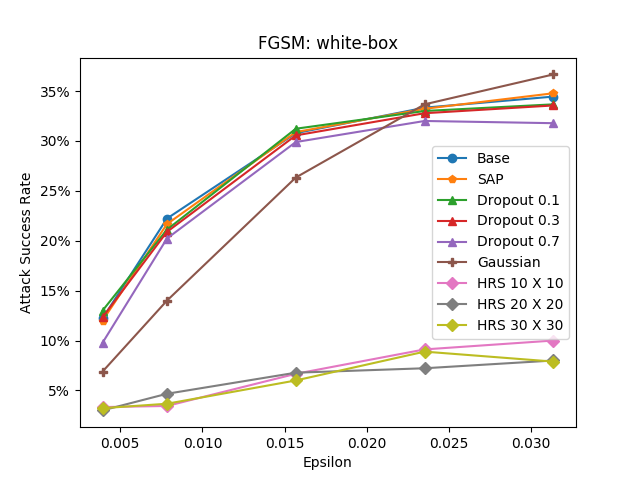}\\
\end{minipage}
}
\caption{Attack success rate of FGSM on different defenses.}
    \label{Figure:CIFAR_FGSM}
\end{figure}

\begin{figure}[ht]
    \centering
    \vspace{-15mm}
    \subfigure{
\centering 
\begin{minipage}[b]{0.45\textwidth}
\centering  
\includegraphics[width=1\textwidth]{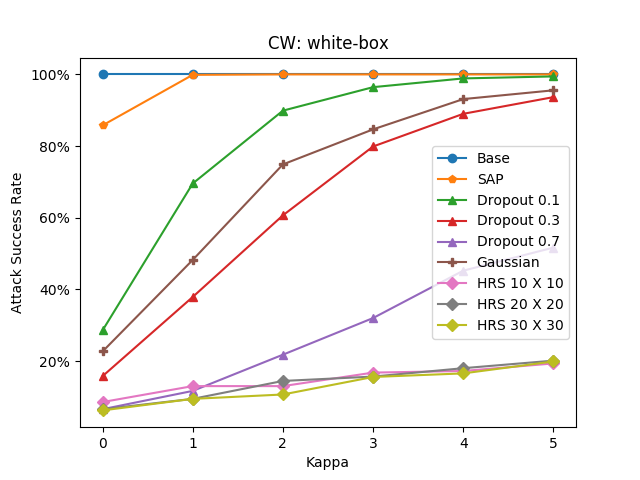} \\
\end{minipage}
}
    \subfigure{
\centering 
\begin{minipage}[b]{0.45\textwidth}
\centering  
\includegraphics[width=1\textwidth]{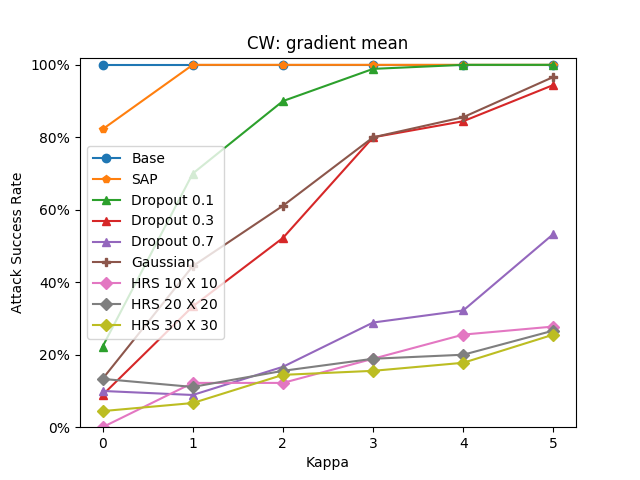}
\end{minipage}
}
    \subfigure{
\centering 
\begin{minipage}[b]{0.45\textwidth}
\centering  
\includegraphics[width=1\textwidth]{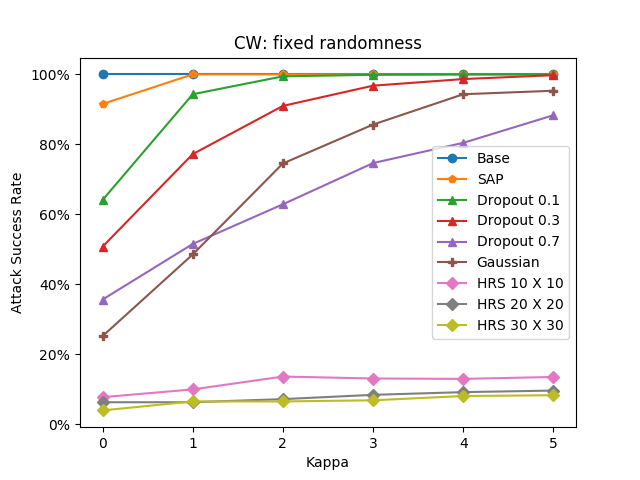}\\
\end{minipage}
}
\caption{Attack success rate of CW on different defenses.}
    \label{Figure:CIFAR_CW}
\end{figure}

\clearpage
\section{Adversarial Reprogramming}
\label{sup_reprogramming}

The number of parameters in the input transformation is crucial factor to achieve high reprogramming accuracy. We applied the original input transformation in \cite{elsayed2018adversarial} to our CIFAR to MNIST reprogramming task but find the reprogramming performance is poor due to the lack of parameters in input transformation. In order to differentiate defense effectiveness under a strong reprogramming setting, we use a local-connected layer as input transformation. The advantage of local-connected layer is that we can easily control the number of parameters by setting different kernel sizes. We found that using a $3\times3$ kernel lead to ~95.07\% accuracy which is similar to reported accuracy in \cite{elsayed2018adversarial}. We show experiment results using other kernel sizes in the following and it is clear that using a larger kernel (a large number of parameters) will lead to higher reprogramming accuracy. However, under all experiment settings we found our proposed defense demonstrate much stronger defense (lower reprogramming accuracy) compared to other defenses.

\clearpage
\section{Supplementary Experiments}
\label{supp_exps}

\subsection{The Effect of Increasing Switching Blocks}
In Table \ref{tab:nb_block} we compare HRS models with 1, 2 and 3 switching blocks. For all models in the comparison, there are 5 channels in each block. Thus the parameter size of these 3 models are the same. It is noted that by increasing the number of switching blocks, the resistance against adversarial attacks can be improved. So the benefit of using more switching blocks is increasing model variation given certain parameter size, and thus improving the defending effectiveness. Yet the improvement is traded with more test accuracy drop. The test accuracy of these three models are 78.78\%, 78.74\% and 74.92\% respectively. Therefore, the number of blocks of a HRS model can be treated as a defense strength controlling factor.

\begin{table}[h]
    \centering
    \caption{Defense Effectiveness (in terms of ASR) of HRS with different number of blocks. The experiment is conducted on CIFAR-10 dataset using CW-PGD attack with different $\epsilon$ bounds.}
    \label{tab:nb_block}
    \begin{tabular}{|c|c|c|c|c|c|}
        \hline
        Strength & 1/255 & 2/255 & 4/255 & 6/255 & 8/255 \\
        \hline
        \hline
        1-block & 5.6\% & 16.4\% & 46.7\% & 76.7\% & 85.5\% \\
        \hline
        2-block & 3.3\% & 10.0\% & 35.6\% & 64.3\% & 81.2\% \\
        \hline
        3-block & 2.9\% & 13.1\% & 33.7\% & 55.2\% & 65.9\% \\
        \hline
    \end{tabular}
\end{table}




\end{document}